\documentclass[preprint,nonatbib]{article}

 \usepackage{neurips_2019}

\usepackage[utf8]{inputenc} % allow utf-8 input
\usepackage[T1]{fontenc}    % use 8-bit T1 fonts
\usepackage{hyperref}       % hyperlinks
\usepackage{url}            % simple URL typesetting
\usepackage{booktabs}       % professional-quality tables
\usepackage{amsfonts}       % blackboard math symbols
\usepackage{nicefrac}       % compact symbols for 1/2, etc.
\usepackage{microtype}      % microtypography
\usepackage{graphicx}
\usepackage[labelfont=bf]{caption}
\usepackage[superscript,nomove]{cite}
\usepackage{color}
\usepackage{wrapfig}
\usepackage{subfig}
\usepackage{graphicx}
\usepackage{subfig}
\usepackage{listings}
\usepackage{colortbl}
\usepackage{xcolor}

\definecolor{pykeyword}{HTML}{37AC4A} % as reported by powerpoint eyedropper
\definecolor{pyoperator}{HTML}{A51DFF} % as reported by powerpoint eyedropper

\lstdefinelanguage{python}{
  morestring=[b]',
  morestring=[b]""",
  morecomment=[l]\#,
  morekeywords={and,as,assert,break,class,continue,def,del,elif,else,except,False,finally,for,from,global,if,import,in,is,lambda,None,nonlocal,not,or,pass,raise,return,True,try,while,with,yield},
  keywordstyle=\color{pykeyword}\bf,
  otherkeywords={|,>>,\&},
  morekeywords=[2]{|,>>,\&},
  keywordstyle=[2]\color{pyoperator},
}

\title{A semi-supervised deep learning algorithm for abnormal EEG identification}

\author{%
	Subhrajit Roy, Kiran Kate, and Martin Hirzel\\
	IBM Research\\
  \textsf{\small roy.subhrajit20@gmail.com, kakate@us.ibm.com, hirzel@us.ibm.com}}
\begin{document}

\maketitle

\begin{abstract}
Systems that can automatically analyze EEG signals can aid
neurologists by reducing heavy workload and delays. However, such
systems need to be first trained using a labeled dataset. While large
corpuses of EEG data exist, a fraction of them are labeled.
Hand-labeling data increases workload for the very neurologists we try
to aid. This paper proposes a semi-supervised learning workflow that
can not only extract meaningful information from large unlabeled EEG
datasets but also make predictions with minimal supervision, using
labeled datasets as small as 5~examples.
\end{abstract}

\begin{figure}[!h]
	\centering
	\includegraphics[width=0.63\textwidth]{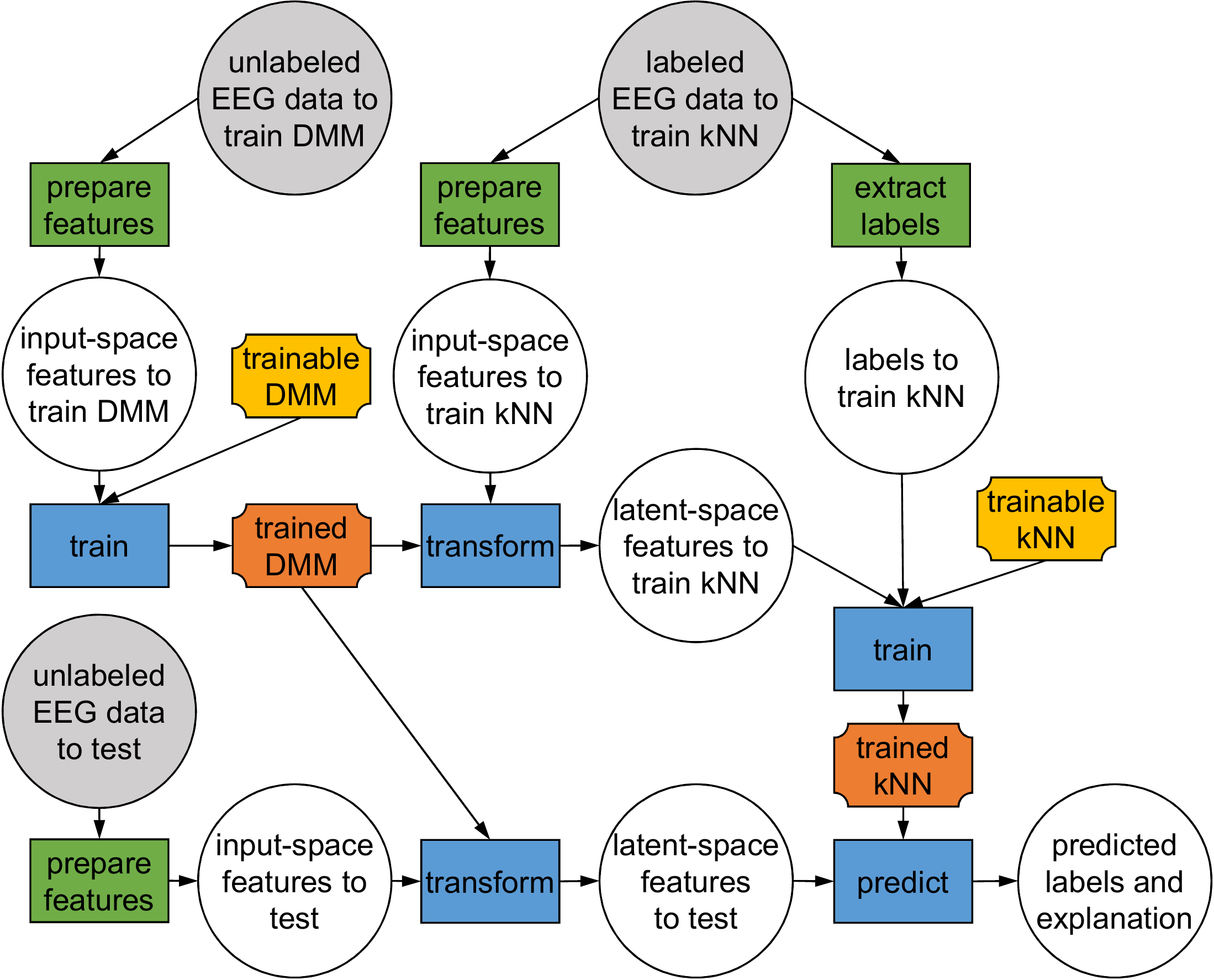}
	\caption{Proposed semi-supervised learning workflow for abnormal EEG identification.}\label{fig1}
	\vspace*{1mm}
\end{figure}

\section{Introduction}\label{sec:intro}

Brain-related disorders such as epilepsy can be diagnosed by analyzing electroencephalograms (EEGs). However, manual analysis of EEG data is time-consuming due to the relatively low availability of expert investigators. Hence, automatic EEG interpretation by machine-learning algorithms has gained popularity recently \cite{ref_eeg1,ref_eeg2,ref_eeg3}. However, typically such algorithms require a large labeled dataset to train on. It is not always possible to obtain such a dataset, since there is a limited number of certified EEG labelers. This paper tackles this problem by proposing a semi-supervised learning workflow for classifying EEGs, comprising an unsupervised learning phase followed by supervised learning (similarly to Kingma et al.'s M1 workflow\cite{kingma_et_al_2014}).
The unsupervised phase trains a Deep Markov Model (DMM)\cite{krishnan_shalit_sontag_2017} to learn non-linear sequential dependencies in EEG signals from a large set of unlabeled EEG data. The supervised phase uses the trained DMM and a small set of labeled EEG data to obtain latent features for training a k-Nearest Neighbors (kNN) algorithm. Using kNN helps explain predictions by returning similar cases. This paper concentrates on one of the first steps in interpreting an EEG session: identifying whether the brain activity of a patient is abnormal or normal. To train and evaluate the proposed system, we use the TUH EEG Abnormal Corpus dataset\cite{tuh_eeg_2015}, which consists of 1,488 abnormal and 1,529 normal labeled EEG sessions. The dataset was reorganized into a training set (1,361 abnormal/1,379 normal) and a test set (127 abnormal/150 normal). 

\section{Methods}

Figure~\ref{fig1} summarizes the complete training and evaluation process.
Neurologists typically classify an EEG session into either normal or abnormal by examining only its initial segment \cite{lopez_2017}. Hence, like L{\'o}pez et al.\cite{lopez_2017}, we extracted only the first minute of each EEG session from the training and test set. Next, we converted the recorded raw EEG signal into the transverse central parietal (TCP)\cite{lopez_2017} montage system for accentuating spike activity. We extracted four standard features (power in the alpha, beta, theta, and delta band) from each second of data. Figure~\ref{fig1} refers to these as input-space features.
The training dataset consists of a large unlabeled set and a small labeled set. 

\begin{figure}
	\centering
	\includegraphics[width=0.38\textwidth]{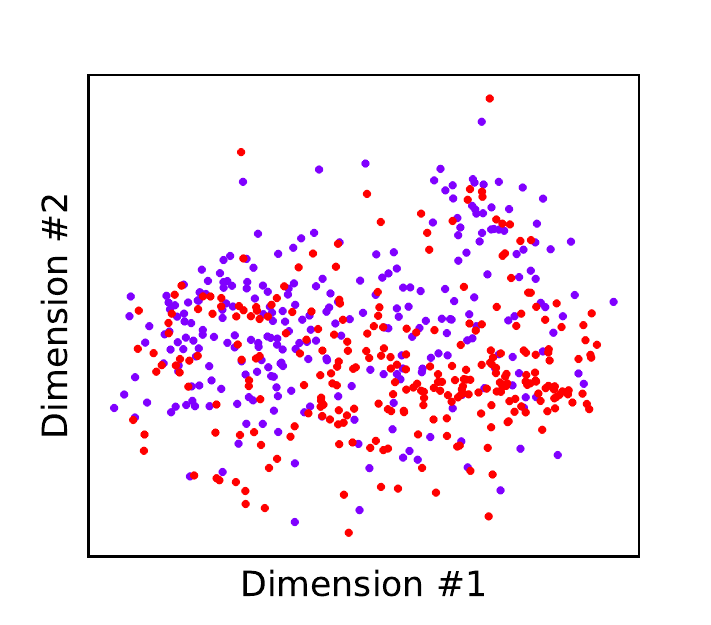}
	\includegraphics[width=0.38\textwidth]{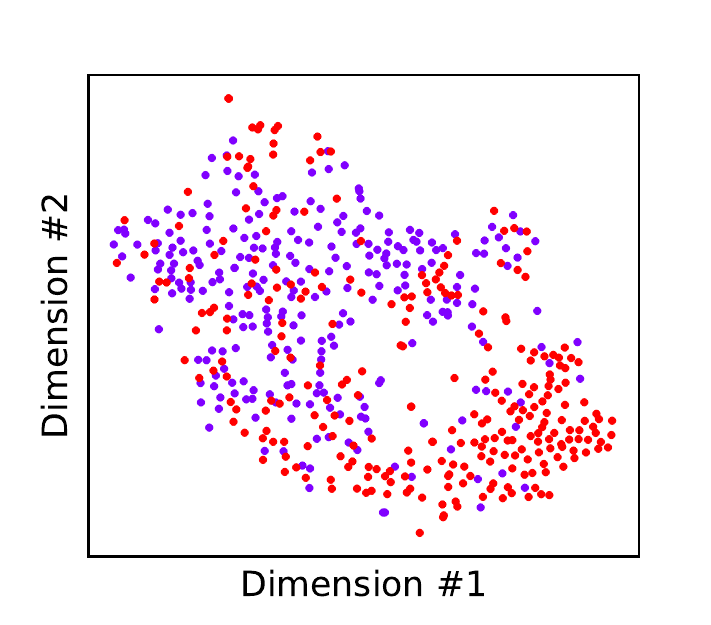}
	\caption{TSNE visualization of the input-space (left) and the latent-space features learned by the DMM (right). Red dots correspond to normal EEGs and blue dots correspond to abnormal EEGs.}\label{fig3}
\end{figure}

We trained a DMM\cite{krishnan_shalit_sontag_2017} on the unlabeled training dataset to model the dynamics of the EEG features over time.  DMM is a high-dimensional, non-conjugate model designed to be fit to large data sets. The number of latent variables in a
sequence depends on the input data. Compared to a Markov model, a DMM
is flexible enough to capture highly non-linear dynamics. This is
because in a DMM, the transition probabilities that govern the
dynamics of the latent variables as well as the emission probabilities
governing how the observations are generated by the latent dynamics
are parameterized by deep neural networks (hence the name Deep Markov
Model). This makes a DMM particularly well-suited for modeling EEG
data. We used Krishnan et al.'s DMM architecture for modeling
temporal dependencies\cite{krishnan_shalit_sontag_2017}.
Figure \ref{fig3} shows the t-distributed stochastic neighbor
embedding (TSNE) visualization of the input space and the features
learned by the DMM, where blue and red dots correspond to normal and
abnormal EEG, respectively. The latter qualitatively depicts that the
DMM learns more discriminative features in the latent space as
compared to the input space.

\begin{figure}
  \centerline{\includegraphics[width=0.8\textwidth]{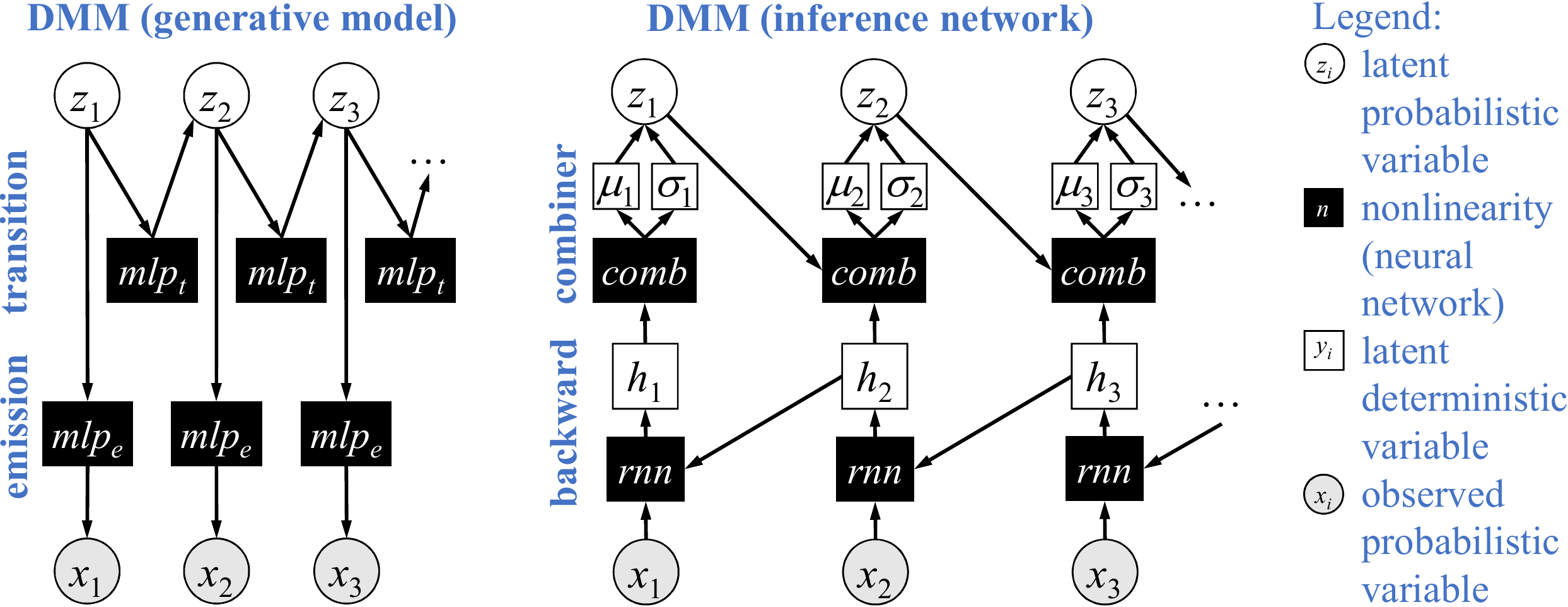}}
  \caption{\label{fig_dmm}Deep Markov Model (DMM) graphical model and guide.}
\end{figure}

Figure~\ref{fig_dmm} shows the the DMM architecture, comprising a
\emph{generative} and an \emph{inference} model.  The \emph{generative
  model} takes a sequence $\vec{z}$ of latents and generates the
corresponding observations~$\vec{x}$. The model's transition and
emission functions are modeled by multi-layered perceptrons
(MLPs). The transition module uses a gated transition function without
being conditioned by the observations, similar to the Markovian
properties of the latents. The \emph{inference network} serves as a
variational guide~\cite{blei_kucukelbir_mcauliffe_2017,baudart_hirzel_mandel_2018}, taking an
observation sequence $\vec{x}$ to propose corresponding
latents~$\vec{z}$. The guide is structured upon the factorization of
the posterior latent distribution. The factorization is followed by
using a backward (right to left) recurrent neural network (RNN), which
outputs a hidden unit $h_i$ for each time step~$i$. A
combiner function uses $h_i$ and $z_{i-1}$ to propose the
approximate latent $z_i$.

We trained the DMM for 50 epochs with a batch size of 32 and learning rate of 0.001. We used the stochastic variational inference strategy and ADAM optimization algorithm. Our implementation is based on the auto-gradient computation framework of the Pyro library \cite{bingham_et_al_2019}. Once the DMM is trained, we use the labeled training samples and extract their corresponding latent-space features from the trained DMM. Next, these latent features are used to train the k-Nearest Neighbors (kNN) algorithm for identifying abnormal EEG sessions.
We used the scikit-learn implementation \cite{buitinck_et_al_2013} of the kNN algorithm.
We chose the kNN algorithm since it offers a certain level of interpretability by explaining its classification decisions to neurologists via examples.

During evaluation, we first pass the test set through the trained DMM
and obtain the features from the latent space. Next, we use the
trained kNN algorithm to obtain the predictions. If requested by
neurologists, the model can also return the nearest neighbors from the
labeled training data as explanation.
Given an unlabeled sample, the \emph{kneighbors} method from the
scikit-learn implementation of kNN returns the
indices of the $k$ labeled samples that are nearest to it in latent
space.  Using those indices, we can retrieve the
corresponding labeled points in input space as a visual explanation
for the prediction. We used the Hyperopt
library\cite{bergstra_et_al_2015} to explore and optimize the
hyperparameters of kNN.

\section{Results and discussion}

\begin{figure}
  \centering
  \subfloat[\label{a}]{%
  \hspace*{-2mm}\includegraphics[width=0.34\textwidth]{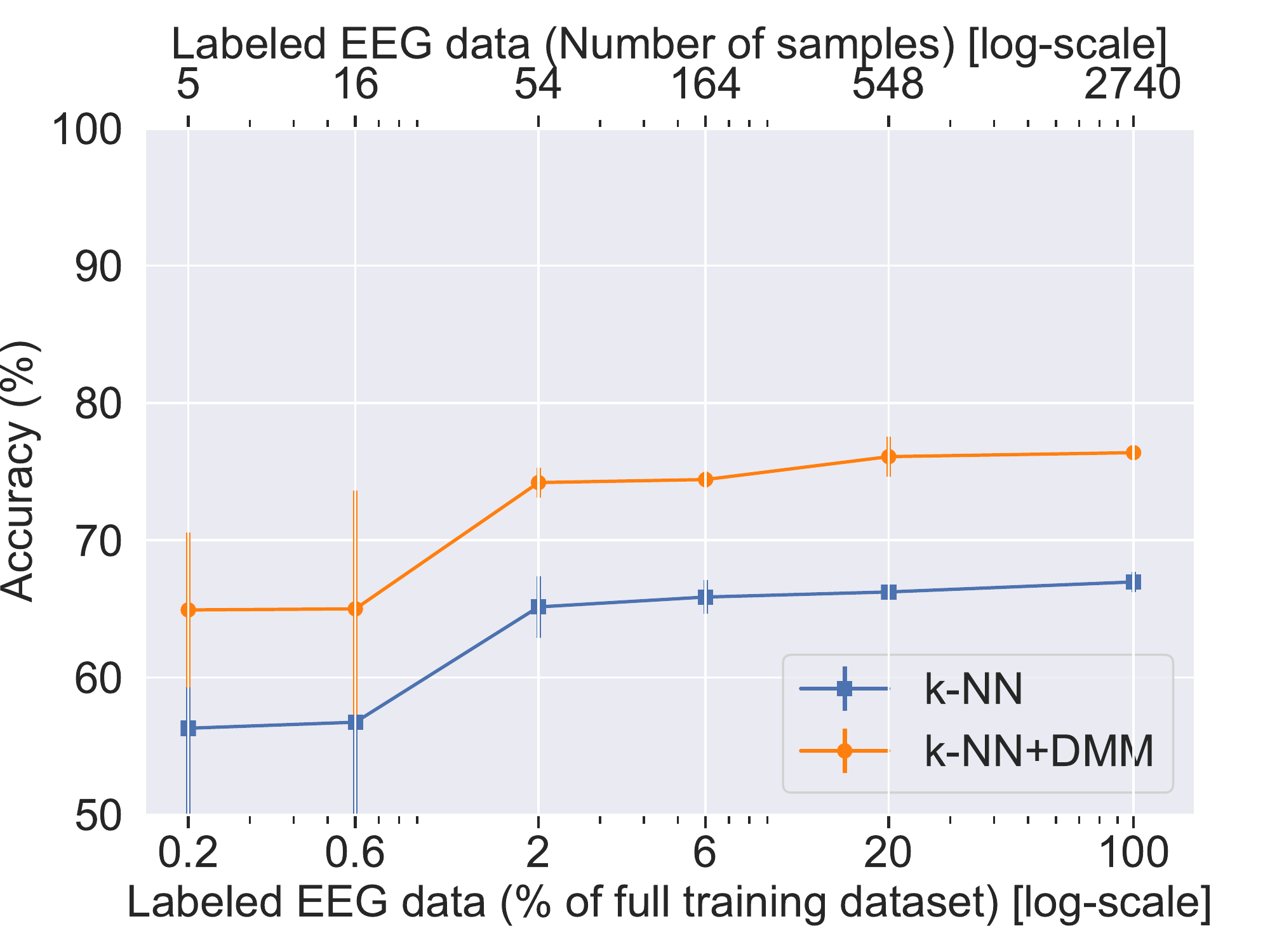}
  }
  \subfloat[\label{b}]{%
  \includegraphics[width=0.34\textwidth]{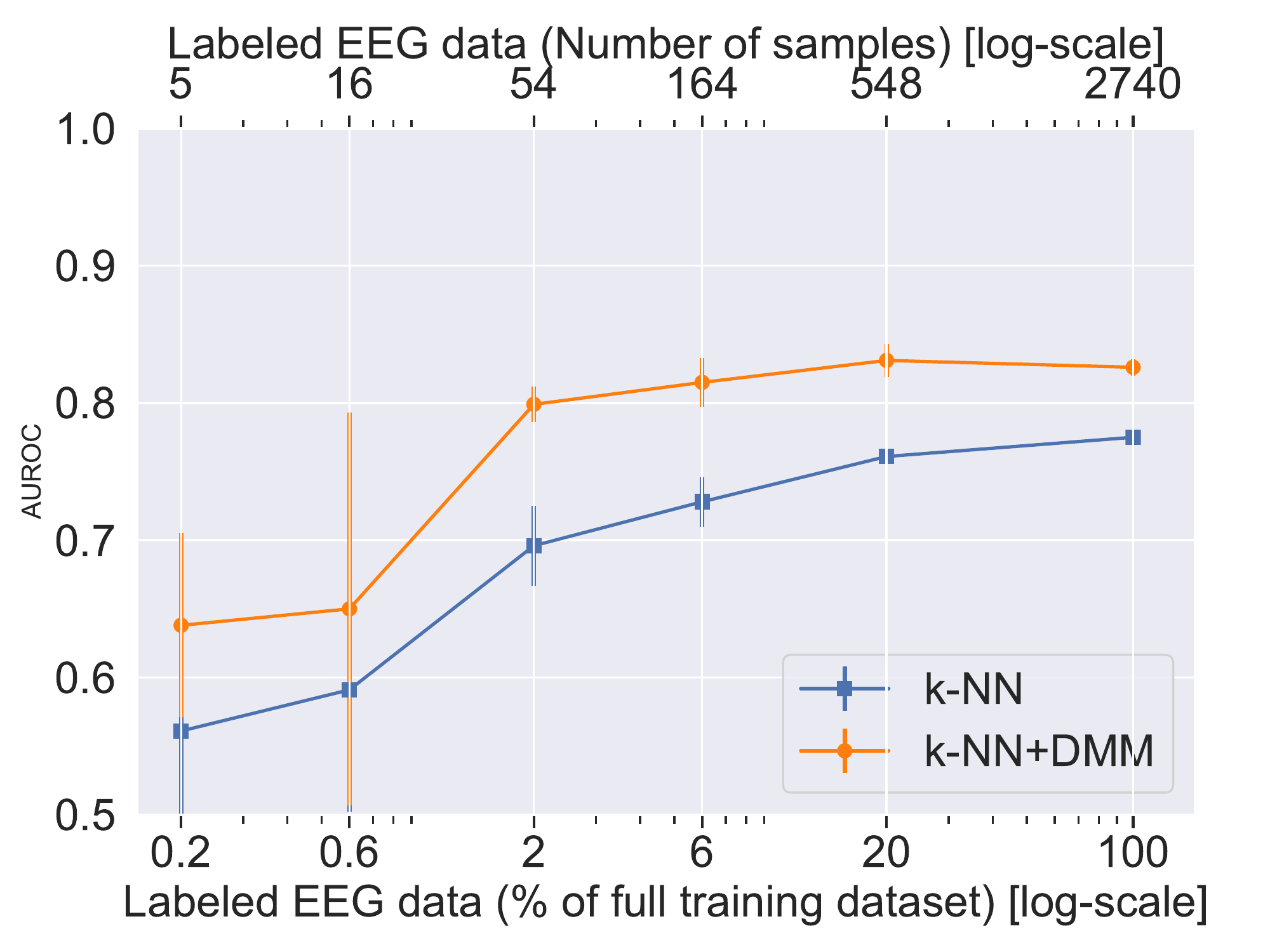}
  }
  \subfloat[\label{c}]{%
  \includegraphics[width=0.34\textwidth]{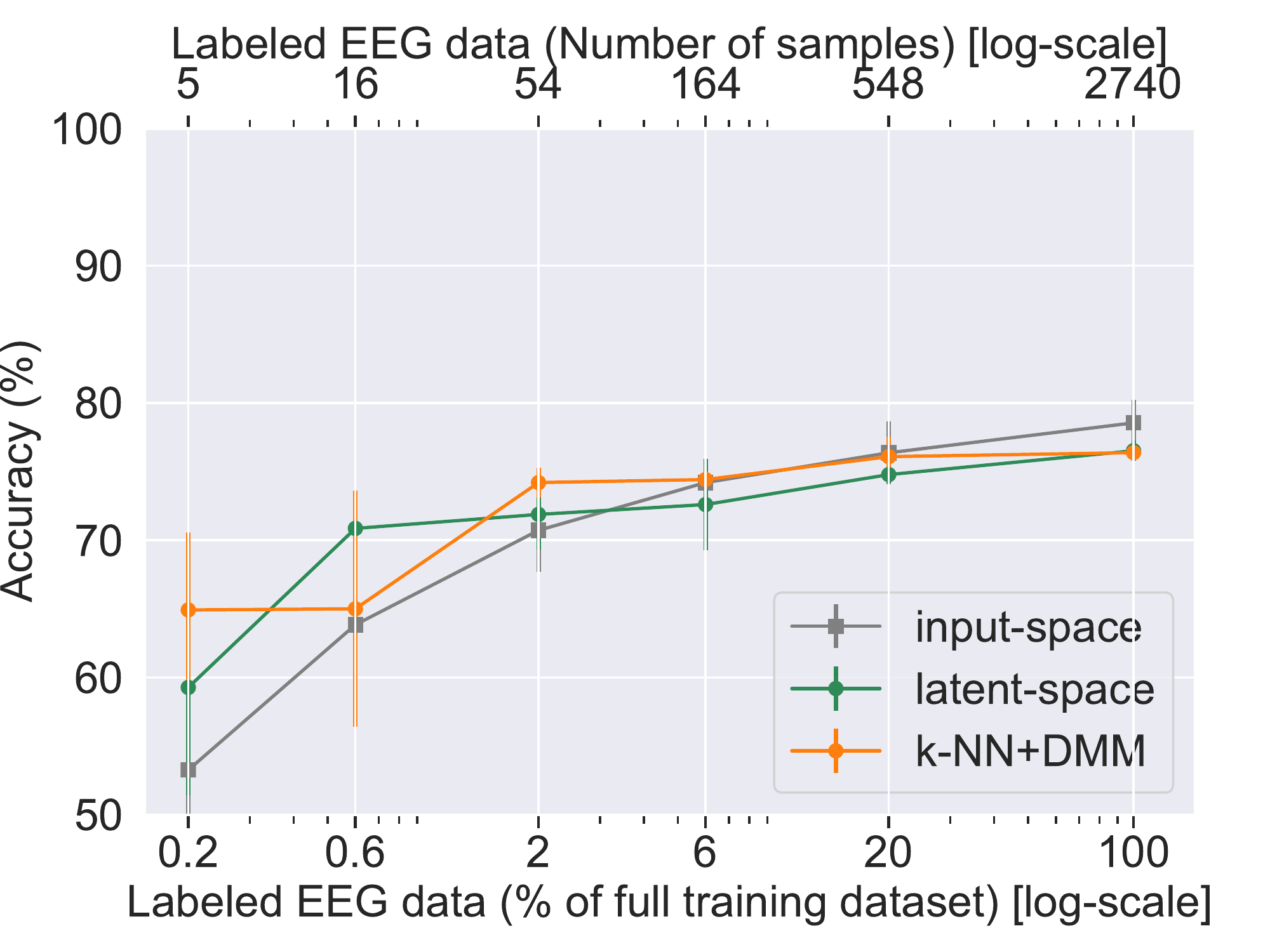}
  }
  \caption{Test accuracy (a, c) and AUROC (b) vs. size of labeled dataset (top horizontal axis: number of 
  samples, bottom axis: percentage of full training data). 
  Each data point shows the average and standard deviation obtained across 5 runs. (c) is the comparison of best classification performance among 11 classifiers
  on input-space features and latent-space features to our approach (k-NN+DMM).}
  \label{fig2}
\end{figure}

To analyze the performance of our proposed workflow, we conduct two
experiments, both using the train/test split discussed in
Section~\ref{sec:intro}. We use the full training set without
labels to train the DMM.  Then, we pick a random stratified subset of
the training set for the supervised phase.  We vary the amount of data
used in the supervised phase for training the classifier and obtain
classification performance on the test dataset.

\begin{figure}
  \centering
  \subfloat[\label{a}]{\hspace*{-2mm}\includegraphics[width=0.34\textwidth]{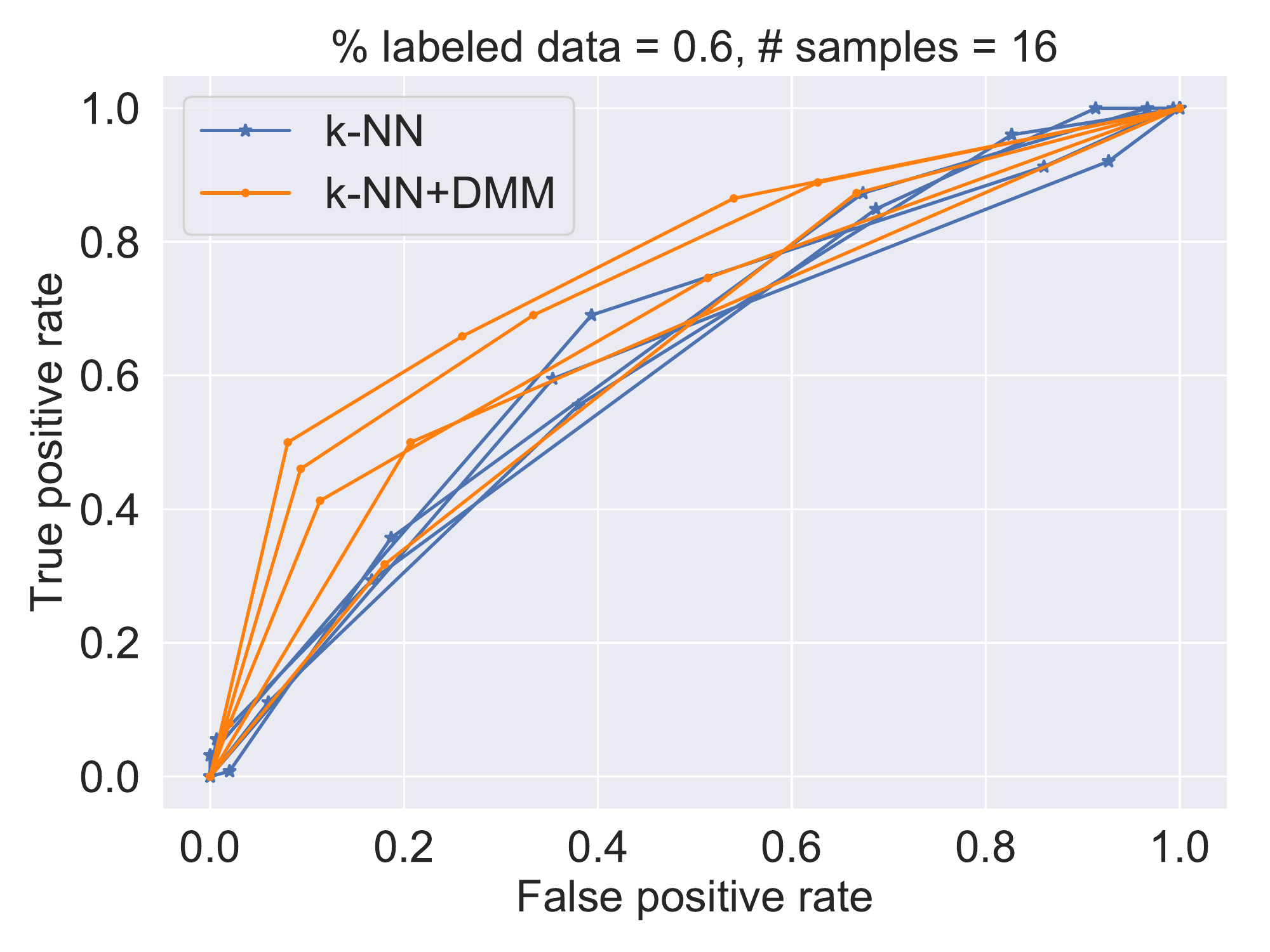}}
  \subfloat[\label{b}]{\includegraphics[width=0.34\textwidth]{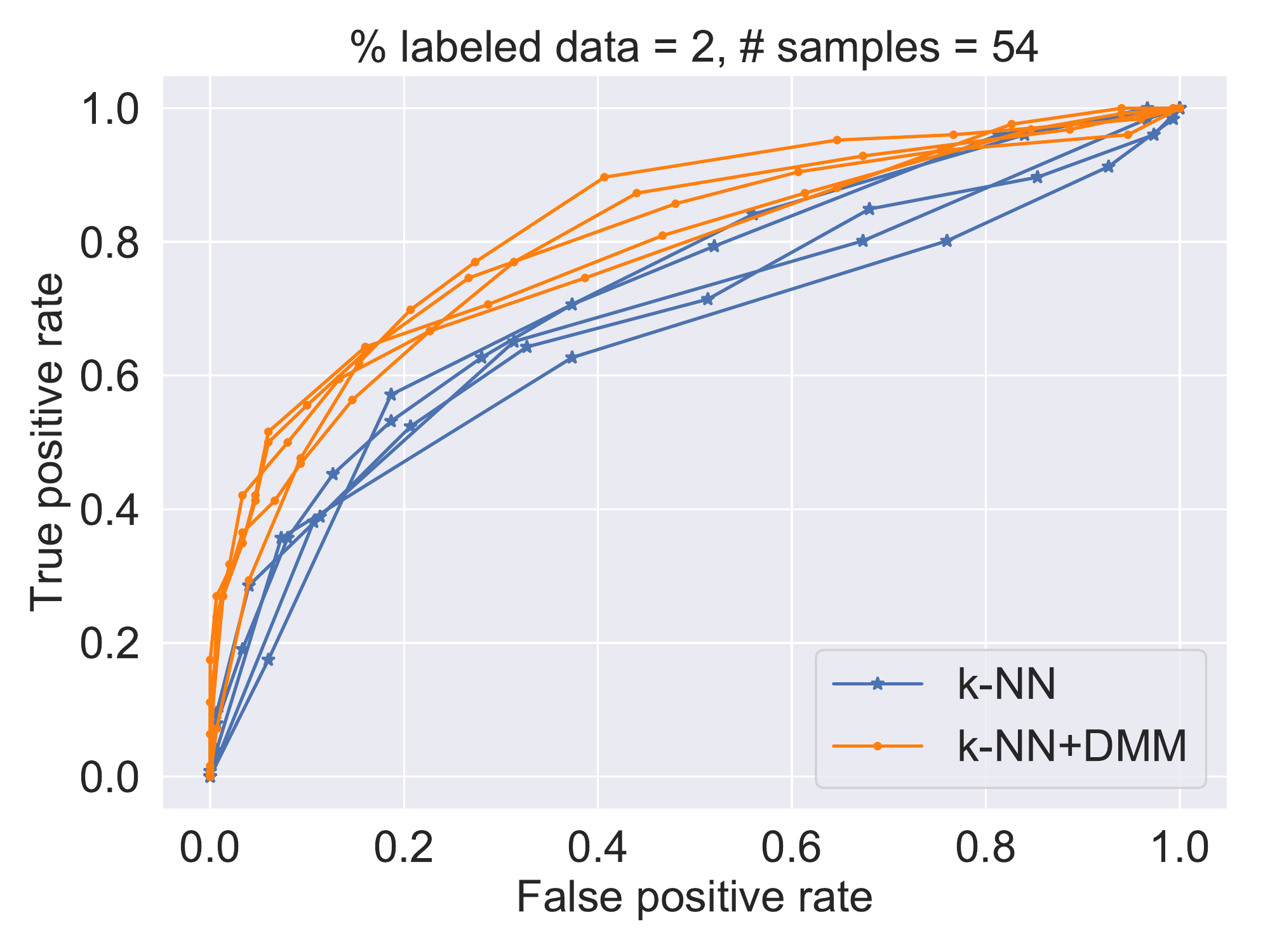}}
  \subfloat[\label{c}]{\includegraphics[width=0.34\textwidth]{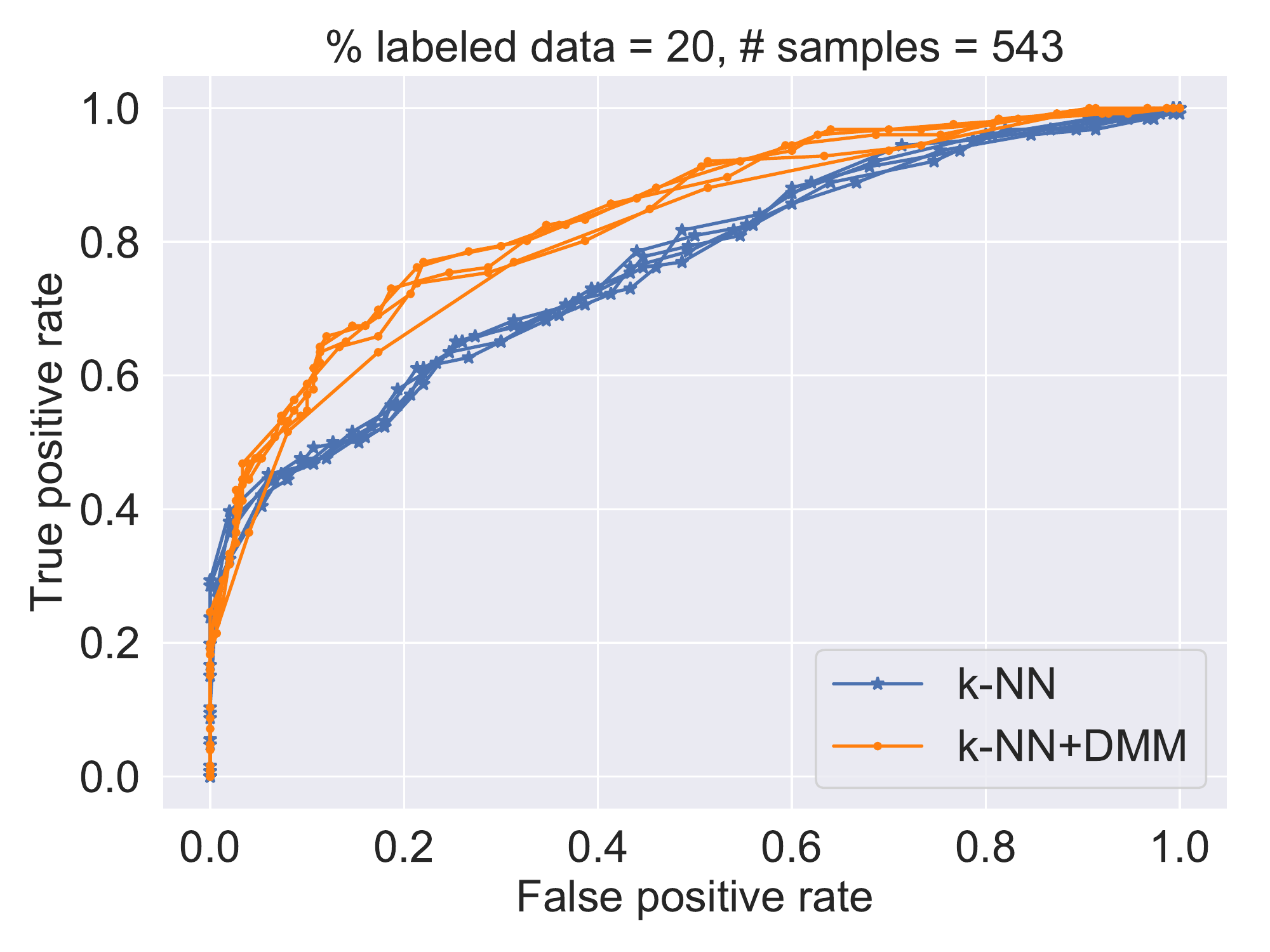}}
  \caption{ROC curves of 5 runs for different sizes of the labeled dataset.}\label{roc_curves}
\end{figure}

Our first experiment reports classification accuracies and AUROC on the test dataset when kNN is used as a classifier as proposed
in our system.
These results, reported in Figure \ref{fig2}\subref{a} and \ref{fig2}\subref{b}, 
show that our system can achieve reasonable performance even at 
low amounts of labeled EEG data and the performance gets better with more labels. 
At 20\% labeled data, our model reaches a similar classification accuracy to that reported by L{\'o}pez et al.\cite{lopez_2017}, 
who used the same dataset, a similar preprocessing technique, and trained on 100\% of the data. 
Moreover, Figure \ref{fig2} shows that if, instead of training the kNN on the latent features extracted from the trained DMM, 
we directly train it on the input space, it performs worse. 
This shows that the DMM is learning meaningful representations during the unsupervised training process. 
To further demonstrate the advantages of the proposed method, we dive deeper and show the ROC curves obtained at 
different levels of the size of the labeled dataset for the 5 runs in Figure \ref{roc_curves}.
The ROC curves confirm the AUROC results and demonstrate that our method
performs better across the spectrum of true/false positive trade-offs.
%% It is evident from the curves that in general our method performs
%% better in distinguishing between the normal and abnormal EEG
%% sessions, thereby leading to a higher value of AUROC in Figure
%% \ref{fig2}.

Our second experiment tests the suitability of kNN as a classifier. While kNN is an important choice for explainability,
we would want to make sure that classification performance is not affected negatively. We performed combined algorithm selection
and hyperparameter tuning (CASH) on 11 popular classifiers: 9 from scikit-learn\cite{buitinck_et_al_2013} along with the XGBoost Classifier\cite{Chen:XGB} and the
LightGBM Classifier\cite{LightGBM} as shown in Figure \ref{fig:CASH}.
We used a Python library Lale\cite{arxiv19-lale} that simplifies CASH using Hyperopt. We used the same CASH budget of 150 Hyperopt
trials as the previous experiment. Figure~\ref{fig:CASH} shows a code snippet of this experiment. Figure \ref{fig2}\subref{c} shows that for low amounts of labeled EEG data, our proposed approach still outperforms even the best 
classifier trained on input-space features. Even in the case with 100\% labeled data, we gain interpretability at slightly worse
performance. Interestingly, kNN on the latent space performs 
better than the best classifier for the latent space found using CASH. While kNN was among the 11 classifiers used, with limited
budget and other classifiers in the mix, Hyperopt could not tune it to the same extent as in the first experiment.
 
\begin{figure}[!t]
  \begin{lstlisting}[language=python,basicstyle=\scriptsize\sf]    
    from lale.lib.sklearn import GaussianNB, GradientBoostingClassifier as GradBoost, KNeighborsClassifier as KNN, \
                               RandomForestClassifier as RF, ExtraTreesClassifier, QuadraticDiscriminantAnalysis as QDA, \
                               PassiveAggressiveClassifier as PAC, DecisionTreeClassifier as DT, LogisticRegression as LR
    from lale.lib.xgboost import XGBClassifier as XGB
    from lale.lib.lightgbm import LGBMClassifier as LGBM
    from lale.lib.lale import HyperoptClassifier
    planned_pipeline = GaussianNB | GradBoost | KNN | RF | ExtraTreesClassifier | QDA | PAC | DT | LR | XGB | LGBM
    clf = HyperoptClassifier(planned_pipeline, cv = 3, max_evals = 150)
    trained_clf = clf.fit(X_train, y_train)
    predictions = trained_clf.predict(X_test)
    accuracy = accuracy_score(y_test, predictions)

  \end{lstlisting}
  \caption{\label{fig:CASH}Code for CASH experiment discussed in Section 3.}
  \end{figure}

\section{Conclusion}
We propose a semi-supervised learning workflow for automated abnormal
EEG identification. In hospitals, while large volumes of EEG data
exist, typically they are not used to design machine learning systems
due to the absence of annotations. Since this data can only be
reviewed by certified investigators, the amount of annotated data is
bounded by the time these clinicians have. Our solution addresses this
issue by using less annotated data while also extracting relevant
features from the entire unlabeled corpus. We envision that the
proposed workflow might be applicable to other time-series datasets
which we will explore in future.

\bibliographystyle{unsrt}
\bibliography{bibfile}

\end{document}